\documentclass[runningheads,dvipsnames]{llncs}
\usepackage{graphicx}

\usepackage{amsmath,amssymb} 
\usepackage{color}
\usepackage{multirow}
\usepackage{hhline}
\usepackage{makecell}
\usepackage{threeparttable}
\usepackage{floatrow}
\usepackage{comment}
\usepackage{wrapfig}
\usepackage{printlen} 
\usepackage{hyperref}

\usepackage{arydshln}
\renewcommand{\arraystretch}{1}

\makeatletter
\newcommand{\printfnsymbol}[1]{%
        \textsuperscript{\@fnsymbol{#1}}%
}
\makeatother

\usepackage{epsfig}
\usepackage{graphicx}
\usepackage{amsmath}
\usepackage{amssymb}
\usepackage{algorithm}
\usepackage{algorithmic}
\usepackage[dvipsnames]{xcolor}
\usepackage{comment}

\usepackage{rotating}
\usepackage{diagbox}
\usepackage{balance}

\usepackage[utf8]{inputenc} 
\usepackage[T1]{fontenc}    
\usepackage{url}            
\usepackage{booktabs}       
\usepackage{amsfonts}       
\usepackage{nicefrac}       
\usepackage{microtype}      
\usepackage{graphicx}       

\usepackage{paralist}
\usepackage{xspace}
\usepackage{comment}
\usepackage{subfig}

\hypersetup{pagebackref=true,breaklinks=true,letterpaper=true,colorlinks,bookmarks=false}

\newcommand{\myparagraph}[1]{\vspace{0.1em}\noindent\textbf{#1}}

\newcommand{\figureTeaserCompare}
{
\begin{figure*}[!h]
\centering
    \includegraphics[width=0.95\linewidth]{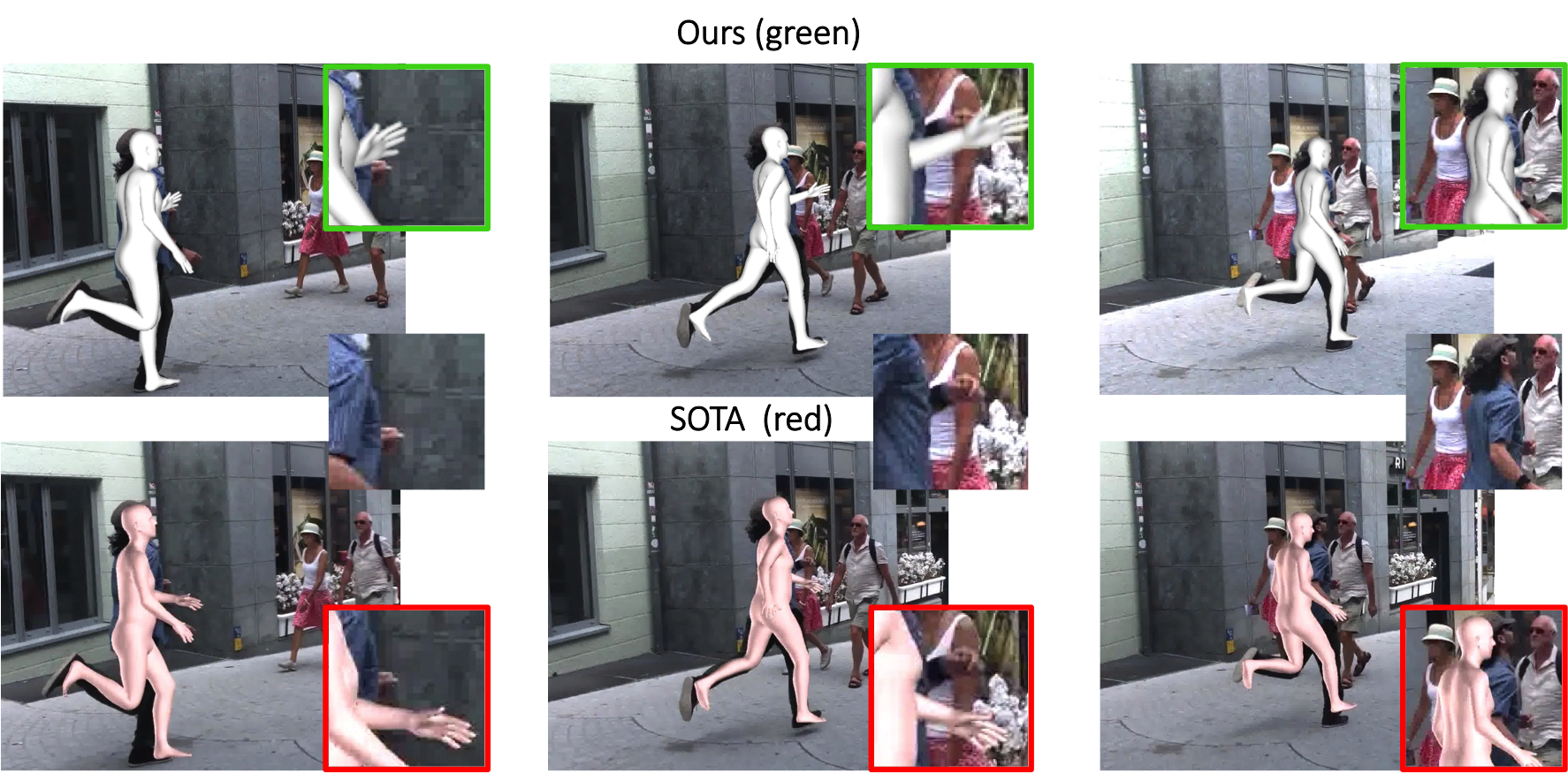}
    \caption{ \textbf{Fitting by Learned Gradient Descent:} We propose a gradient-based iterative optimization technique that combines the refinement capabilities of optimization techniques with the speed and robustness of deep-learning. The approach achieves state-of-the art performance in the most challenging in-the-wild setting, despite not having seen any image data at training time. Compared to the recent SoA regression-based method \cite{kolotouros2019learning} (bottom), \emph{ours} (top) can register fine details such as the lower extremities more precisely and is more robust to visual clutter, such as the pedestrians in the background.   
    }
    \label{fig:teaser}
\end{figure*}
}

\newcommand{\figureInference}
{
\begin{figure*}[t]
\centering
    \includegraphics[width=0.9\linewidth]{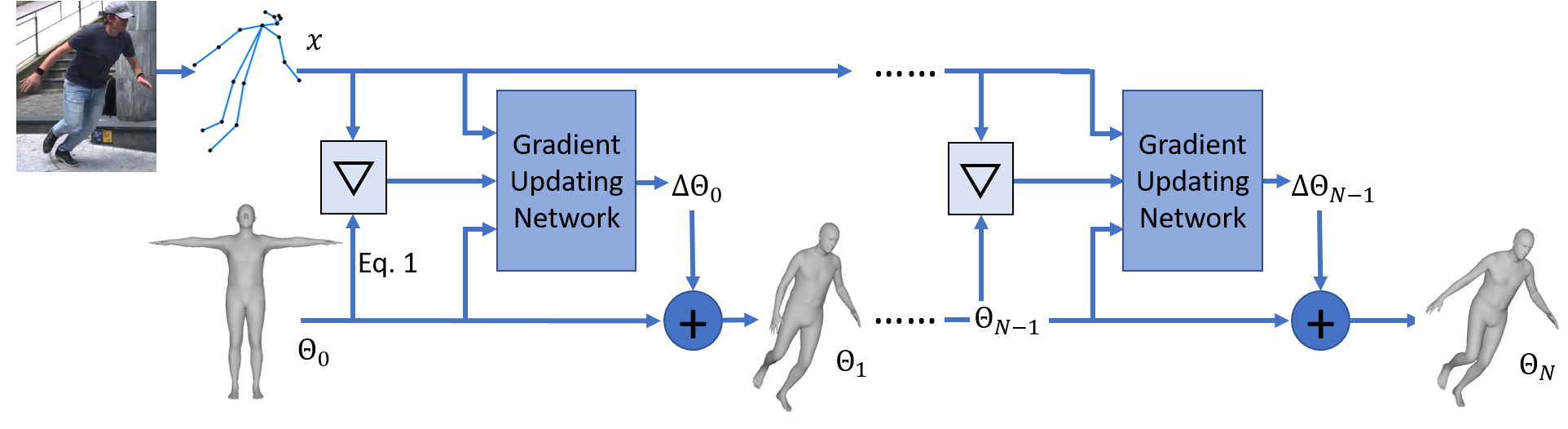}
    \caption{ \textbf{Inference pipeline.} Given a human image, the target 2D keypoints $x$ are obtained from a joint detector. $\Theta_0$ is initialized with zero values. After re-projection via Eq. \ref{equ:projection}, the error between predicted and observed measurements $\mathcal{L}(\Theta_0)$ is calculated, followed by computation of the partial derivative $\frac{\partial \mathcal{L}(\Theta_0) }{\partial \Theta_0}$, denoted by $\nabla $.
Together with the current state of $\Theta_0$ and target $x$, the gradient $\nabla $ is passed into the Gradient Updating Network to obtain the update term $\Delta \Theta_0$. Adding $\Delta \Theta_0$ to $\Theta_0$, yields $\Theta_1$. The whole process will continue until the last iteration N to attain an estimate $\Theta_N$.
    }
    \label{fig:inference}
\end{figure*}
}

\newcommand{\figureNetwork}
{
\begin{figure*}[!h]
\centering
    \includegraphics[width=1\linewidth]{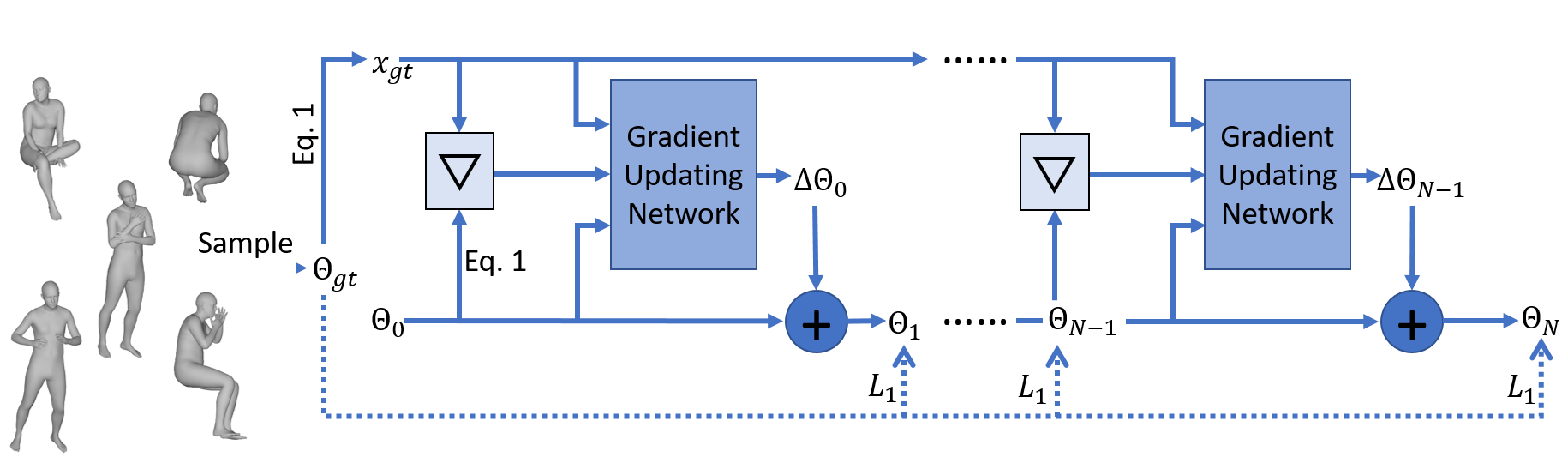}
    \caption{ \textbf{Training scheme.} For training, the \textit{only} source of data we use is a pool
of 3D meshes of human bodies of varying shape and pose. 
To generate a batch of training samples, we randomly sample the pose $\theta_{gt}$ and shape $\beta_{gt}$ pair from the dataset. For each pair of  $\theta_{gt}$ and $\beta_{gt}$, A camera poses ($s_{gt},R_{gt},t_{gt}$ ) is randomly sampled within feasible range. For the sample $\Theta_{gt}$, its corresponding ground-truth 2D keypoints $x_{gt}$ is obtained via re-projection Eq. \ref{equ:projection}. $\Theta_0$ is initialized with zero values. After applying the re-projection Eq. \ref{equ:projection}, the partial gradient $\nabla_0$ wrt $\Theta_0$ is calculated based on the re-projection loss. Once the gradient update term $\Delta \Theta_0$ is obtained from Gradient Updating Network, by adding to $\Theta_0$, $\Theta$ is updated to $\Theta_1$. The whole process will continue till the last iteration N to get the $\Theta_N$. The training loss is only based on the error between true $\Theta_{gt}$ and estimated $\Theta_n$.
    }
    \label{fig:pipeline}
\end{figure*}
}

\newcommand{\figureCollage}
{
\begin{figure*}[!h]
\centering
    \includegraphics[width=1\linewidth]{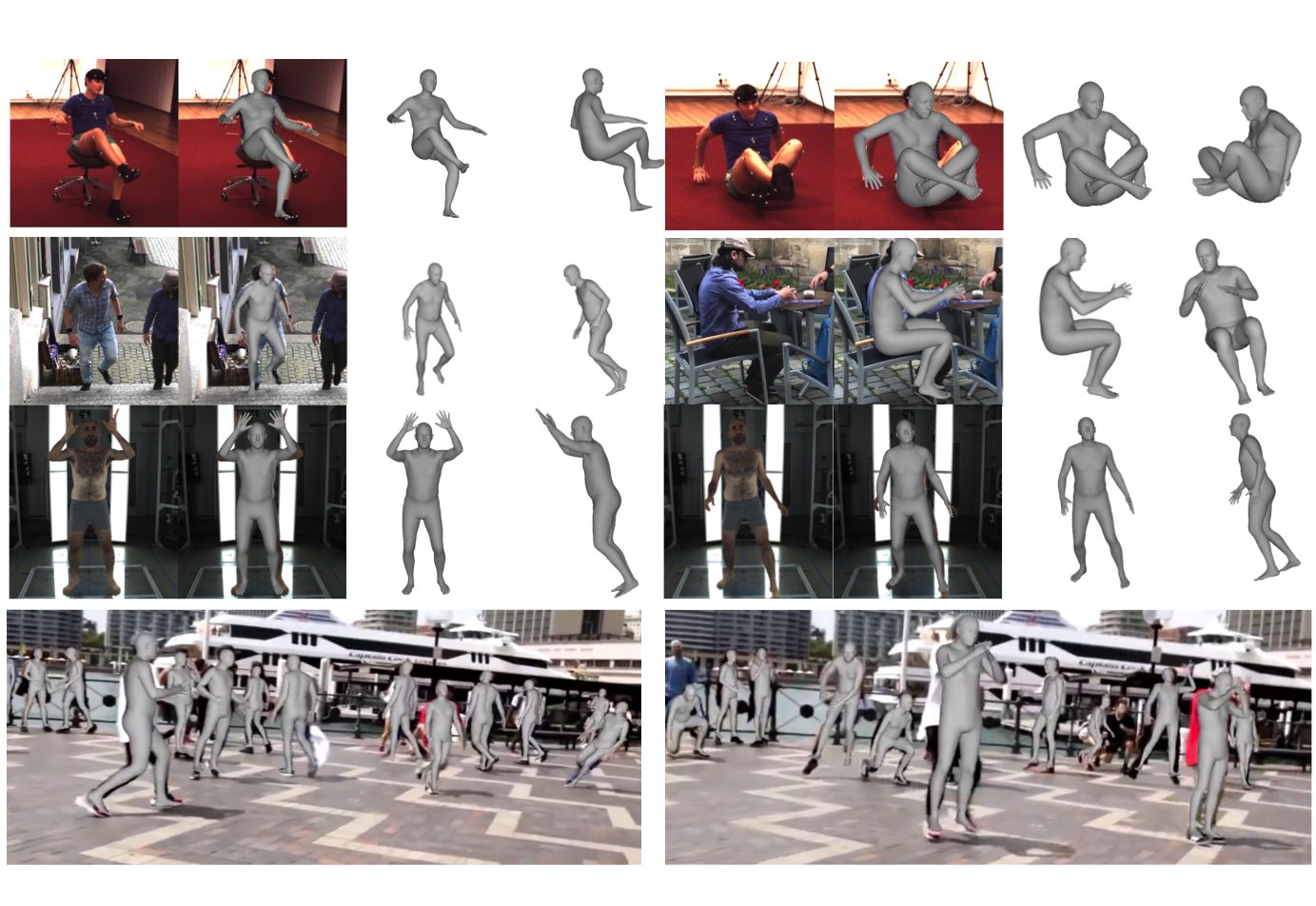}
    \caption{ \textbf{Qualitative results.} Human 3.6M (row 1), 3DPW (row 2),  and EHF (row 3). The last row is the fitting results for a dancing video from the internet. We can see our algorithm generalizes well on random input sources.
    }
    \label{fig:collage}
\end{figure*}
}

\newcommand{\figureComparisonSMPL}
{
\begin{figure}[!h]
\centering
    \includegraphics[width=0.98\linewidth]{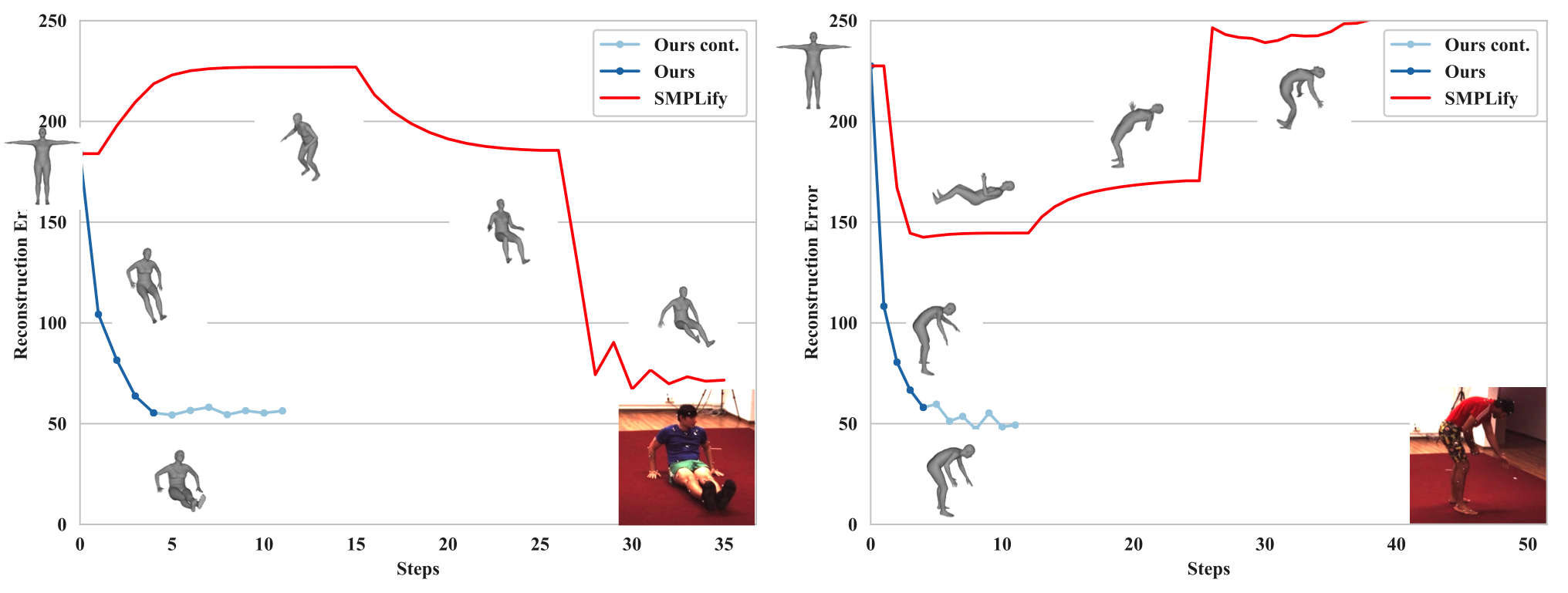}
    \caption{\textbf{Reconstruction error as function of iteration count.} Two examples of fitting progression on the H3.6m test set. Compared to SMPLify (red), our inference converges quicker and is more stable. SMPLify either converges slowly or fails to avoid local minima.  Insets show intermediate pose configurations.
    }
    \label{fig:comparison_smpl}
\end{figure}
}

\newcommand{\figureCompare}
{
\begin{figure*}[!h]
\centering
    \includegraphics[width=1\linewidth]{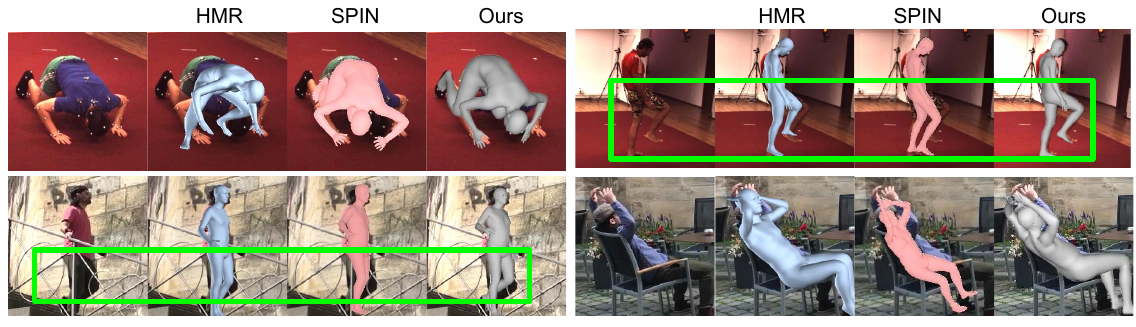}
    \caption{ \textbf{Qualitative comparison with other methods. } Our results align better with the 2D observations thanks to our iterative update scheme.
    }
    \label{fig:compare_new}
\end{figure*}
}

\begin{document}

\pagestyle{headings}
\mainmatter
\def\ECCVSubNumber{100}  

\title{Human Body Model Fitting by Learned Gradient Descent} 

\author{Jie Song\inst{1}\thanks{Equal contribution.} \and
Xu Chen\inst{1,2}\printfnsymbol{1} \and
Otmar Hilliges \inst{1}}
\authorrunning{J. Song et al.}
\institute{ETH Z\"{u}rich \and
Max Planck ETH Center for Learning Systems 
}
\maketitle


\begin{abstract}
We propose a novel algorithm for the fitting of 3D human shape to images. 
Combining the accuracy and refinement capabilities of iterative gradient-based optimization techniques with the robustness of deep neural networks, we propose a gradient descent algorithm that leverages a neural network to predict the parameter update rule for each iteration. This per-parameter and state-aware update guides the optimizer towards a good solution in very few steps, converging in typically few steps. During training our approach only requires MoCap data of human poses, parametrized via SMPL. From this data the network learns a subspace of valid poses and shapes in which optimization is performed much more efficiently. The approach does not require any hard to acquire image-to-3D correspondences.  At test time we only optimize the 2D joint re-projection error without the need for any further priors or regularization terms. We show empirically that this algorithm is fast (avg. 120ms convergence), robust to initialization and dataset, and achieves state-of-the-art results on public evaluation datasets including the challenging 3DPW in-the-wild benchmark (improvement over SMPLify ($45\%$) and also approaches using image-to-3D correspondences). 
\end{abstract}

\section{Introduction}

Recovering the 3D human pose and its shape from a single image is a long standing problem in computer vision
with many downstream applications.
To solve the problem, one has to reconstruct the parameters that characterize human pose and shape from indirect, low-dimensional image observations. Thus, this falls into the category of inverse problems which are generally ill-posed \cite{hadamard1902problemes}.

\figureTeaserCompare
In recent years, the computer vision community has wholeheartedly embraced deep-learning based approaches to such problems. For the case of human shape recovery from monocular images, deep neural networks have been successfully leveraged to regress the parameters of a generative human body model, such as SMPL~\cite{loper2015smpl}, directly from pixel inputs \cite{kanazawa2018end,omran2018neural,tung2017self,kolotouros2019learning}. 
Since it is hard to accurately annotate 3D shapes, very little training data for direct 3D supervision exists (the largest in-the-wild dataset consists of only 60 short sequences \cite{von2018recovering}). Hence, many approaches like \cite{kanazawa2018end} attempt to leverage large-scale 2D keypoint annotations of in-the-wild images via proxy-objectives such as minimizing the distance between ground-truth 2D joint positions and the re-projection of the network predictions. 

Recent work \cite{kolotouros2019learning} proposes to deploy iterative optimization in-the-loop to generate 3D pseudo-labels for further training.  While improving over direct regression, the scalability of this approach is limited since the iterative optimization has to be run until convergence for each training sample. Furthermore, the accuracy of these pseudo-labels remains bounded by the existing optimization method.

In contrast, iterative gradient descent-based optimization methods search for the unknown parameters of the model that best match the available measurements. In the case of human shape recovery, this also often involves minimizing the 2D re-projection error of the rendered human shape with learned priors \cite{bogo2016keep,guan2009estimating,lassner2017unite,pavlakos2019expressive,joo2018total}. These approaches suffer from the non-linear, non-convex, and large-scale nature of the inversion. In consequence, the optimization process tends to be slow and finding good solutions remains very challenging. One reason can be seen in the reliance on multiple regularization and prior terms which have to be traded-off against each other, leading to many possible sub-optimal minima. Furthermore, improving results typically requires domain knowledge and incorporation of heuristics which impact generality.
Despite these difficulties, iterative optimization has many appealing properties such as: 1) not requiring any images with 3D annotation for training, 2) better registration of details to 2D observations compared to end-to-end regression methods via iterative refinement, and 3) no overfitting to a specific dataset. A well tuned optimization algorithm should perform consistently well on different datasets and domains.

In this paper, we take motivation from the optimization and learning communities \cite{adler2017solving,adler2018learned,flynn2019deepview,andrychowicz2016learning} and propose a learning-based update rule for a gradient-based iterative algorithm to fit a human model to 2D. The proposed technique combines the accurate refinement capabilities of optimization techniques with the robustness of learning-based methods when training data is well distributed. Our goal is to make iterative optimization competitive when compared to state-of-the-art regression methods that rely on large amounts of annotated image data (Fig. \ref{fig:teaser}). 

More specifically, we replace the hand-crafted gradient descent update rule with a deep network that has been trained to predict per-parameter, state dependent parameter updates. 
We show that this allows for very efficient optimization. Intuitively, our approach can be seen as joint learning of i) a model prior, ii) regularization terms and iii) gradient prediction. 
That is, the network learns to generate parameter updates that allow the fitting algorithm to stay on
the manifold of natural poses and shape as well as to recover from local minima.
Hence, the optimizer can take larger, parameter-specific steps compared to standard gradient descent, leading to convergence in just a few iterations (typically $\leq 4$ which on the same hardware equates to a speed-up of 500x compared to SMPLify \cite{bogo2016keep}, see Figure \ref{fig:comparison_smpl}). Importantly, at training time the method only requires a dataset of human poses and shapes (e.g., AMASS \cite{AMASS:2019}) and does not use images during training. At inference, the algorithm optimizes the 2D re-projection error directly and does not require any further priors or regularization in order to converge. 
In summary we contribute: 
\begin{itemize}[\textbullet]
    \item A novel iterative algorithm to fit the parameters of a human model to 2D observations via learned gradient descent.
    \item A data efficient way to learn the gradient mapping network, requiring only 3D Mocap data and no image-to-pose correspondences.
    \item Empirical evidence that demonstrates the method is fast, accurate and robust, achieving state-of-the-art results, especially on the most challenging in-the-wild setting (i.e., 3DPW dataset \cite{von2018recovering}).  
\end{itemize}

\section{Related Work}
Our work is related to a large body of research in optimization, machine learning and vision-based pose estimation. Here we briefly review the most related works in human shape estimation and those at the intersection of learning and iterative gradient based optimization.
 
\myparagraph{Human Shape Recovery from Natural Images.}\\
Deep neural networks have significantly advanced skeleton-based 3D human pose estimation from single images \cite{martinez2017simple,mehta2017vnect,zhou2016sparseness,sun2018integral}. In order to obtain more fine-grained representations of the human body, parametric body models such as SCAPE \cite{anguelov2005scape} or SMPL \cite{loper2015smpl} have been introduced to capture the 3D body pose (the skeleton) and its shape (the surface). More expressive models, including hands, feet and face, have recently been proposed  \cite{pavlakos2019expressive,xiang2019monocular}.
Iterative optimization-based approaches have been leveraged for model-based human pose estimation. Early works in the area \cite{guan2009estimating,sigal2008combined,hasler2010multilinear} proposed to estimate the parameters of the human model by leveraging silhouettes or 2D keypoints. In these approaches good correspondences are necessary which are sometimes provided via manual user intervention. More recently, the first fully automatic approach, SMPLify, was introduced by Bogo et al. \cite{bogo2016keep}. Applying an off-the-shelf 2D keypoint detector \cite{pishchulin2016deepcut}, SMPLify iteratively fit the SMPL parameters to the detected 2D keypoints and several strong priors were employed to regularize the optimization process. Lassner et al. \cite{lassner2017unite} leveraged hand-curated results from SMPLify to first train a denser keypoint detector and subsequently incorporate silhouette cues into the fitting procedure.  In \cite{pavlakos2019expressive}, a deep variational autoencoder was proposed as replacement for Gaussian mixtures as pose prior. \cite{varol2018bodynet} proposed to fit SMPL on the regressed volumetric representation obtained from deep networks. Generally speaking, the above model-fitting approaches are not real-time and require about one minute per image or longer. The solutions to the optimization problem are also very sensitive to the choice of the initialization and usually strong regularizing assumptions have to be made in such multi-step  optimization pipelines, which can result in difficulty to tune algorithms.

On the other hand, direct parameter regression via neural networks has been explored as alternative means to the problem of 3D human pose and shape estimation \cite{kanazawa2018end,tung2017self,tan2018indirect,omran2018neural,guler2019holopose,varol2017learning,xu2019denserac,zheng2019deephuman}. Given a single RGB image, a deep network is used to regress the human model parameters. Due to the lack of datasets that contain images with full 3D shape ground truth annotations, these methods have focused on alternative supervision signals to guide the training. These include 2D keypoints, silhouettes, or part segmentation masks. However, such approaches still suffer from coarse estimation in terms of image-model alignment. At the same time, acquiring large amounts of data with image to 3D human shape ground truth correspondences is an extremely hard and cost intensive process. Recent work \cite{kolotouros2019learning} proposes to include iterative optimization in the learning loop to automatically augment the dataset via pseudo-labelling. While demonstrating the promise of combining learning- and optimization-based approaches, the scalability of the approach remains bounded by the run-time speed and the accuracy of the existing optimization method itself. For each training iteration, the pseudo ground truth label is obtained by running the SMPLify method, which itself is slow and may get trapped in local minima even with good initialization. 

\myparagraph{Learned Gradient Descent for Inverse Problem.}\\
Recently research in different domains has suggested to interpret iterative optimization algorithms as unrolled neural networks with a set of inference and model parameters that can be learned jointly via back-propagation \cite{adler2017solving,adler2018learned,flynn2019deepview,andrychowicz2016learning}. 
Andrychowicz et al. \cite{andrychowicz2016learning} proposed to leverage learned gradients in the context of classification tasks. In follow-up work \cite{adler2017solving,adler2018learned} the concept has been advanced into a partially learned gradient scheme, applied to a non-linear tomographic inversion problem with simulated data. Most recently, Flynn et al. \cite{flynn2019deepview} introduced a CNN based framework for the task of novel view synthesis based on multi-plane images. Leveraging learned gradients this approach improves performance on challenging scene features such as object boundaries, thin structures, and scenes with high depth complexity. In this work, we adopt a hybrid approach that incorporates learning of the parameter update into an iterative model fitting algorithm. We model human shape generation as an inverse problem to be solved using a learned gradient descent algorithm. At inference time (see Fig. \ref{fig:inference}), this algorithm iteratively computes gradients of the current human model with regard to the input 2D detections. A network takes these partial derivatives together with the current parameter set as input and generates a per-parameter update rule for the human shape model. We empirically demonstrate that the neural network learns to generate parameter updates that allow the optimization method to stay on the manifold of natural poses and shapes and also to take large steps, thus requiring only a few iterations for convergence.

\figureInference

\section{Method}
\myparagraph{Problem Setting.}\\
Our task is to reconstruct the full 3D mesh of human
bodies from 2D observations (e.g. 2D keypoints extracted from images). The 3D human mesh is encoded by the statistical SMPL body model \cite{loper2015smpl}, which is a differentiable function that outputs a triangulated mesh $M(\theta,\beta)$ that takes as input the pose parameters $\theta\in \mathbb{R}^{23\times 3}$ and the shape parameters $\beta\in \mathbb{R}^{10}$. Specifically,  the template body vertices are first conditioned on $\beta$ and $\theta$, then the bones are articulated according to the joint rotations $\theta$ via forward kinematics, and finally the surface is deformed with linear blend skinning to return the body mesh $M\in \mathbb{R}^{N\times 3}$, with $N= 6890$ vertices. Conveniently, the 3D body joints $X\in \mathbb{R}^{k\times 3}$ of the model can be defined as a linear combination of the mesh vertices.  A linear regressor $W$ is usually pre-trained to map the vertices to $k$ joints of interest, defined as $X = W M$.  To re-project 3D joints to 2D, a weak-perspective camera model is employed. The camera model is parameterized by the global rotation $R\in \mathbb{R}^{3\times 3}$ in axis-angle representation, translation $t\in \mathbb{R}^{2}$ and scale  $s\in \mathbb{R}$. Thus the set of parameters that represent the reconstruction of a human body is expressed as a vector $\Theta = \left \{\theta,\beta,R,t,s  \right \}$.  Given $\Theta$, the projection of 3D joints $X$ to 2D is:
{
\begin{align}
\hat{x} = s\Pi(RX(\theta,\beta)) + t,
\label{equ:projection}
\end{align}
}where $\Pi$ is an orthographic projection.

In this paper, we seek to solve the inverse problem associated with Eq. \ref{equ:projection}. That is, we wish to compute a set of parameters $\Theta = \left \{\theta,\beta,R,t,s  \right \}$ that match the observed 2D keypoints $x$.  Since the number of model parameters $\Theta$ is typically larger than the number of 2D measurements, this inverse problem is ill-posed and solving it usually requires additional priors and strong regularizers. Here we seek a method that finds good solutions without requiring such auxilliary measures.
\\

\myparagraph{Iterative Optimization with Explicit Regularization.}\\
In most inverse problems a closed-form map from observations to signal is intractable.
Inverse problems are often solved via iterative optimization by minimization, e.g.:
{
\begin{align}
\underset{\Theta}{\mathrm{argmin}}= \mathcal{L}(\Theta)+ \Phi(\Theta),
\label{equ:overall_loss}
\end{align}
}%
where $\mathcal{L}(\Theta) = L_{reproj}(\hat{x}, x)$ is the data term measuring the agreement between predicted and observed measurements. More specifically, in our context, $\hat{x}$ are the projected 2D joints from SMPL parameters computed by Eq. \ref{equ:projection}. $\Phi(\Theta)$ is a prior term on $\Theta$ to regularize the optimization process via maximum a posteriori. Such non-linear optimization problems can be solved via iterative methods such as stochastic gradient descent. The update rule (with step size 
$\lambda$) is given by:
{
\begin{align}
\Theta_{n+1} = \Theta_n + \lambda[\frac{\partial \mathcal{L}(\Theta_n) }{\partial \Theta_n} +\frac{\partial \Phi(\Theta_n) }{\partial \Theta_n}]
\label{equ:updaterule_old}
\end{align}
}

For example, in the seminal work of SMPLify by Bogo et al. \cite{bogo2016keep}  $\Phi(\Theta)$ corresponds to several prior terms defined on both $\theta$ and $\beta$. The priors are usually seperately pre-trained on fitted MoCap data and serve the purpose of penalizing implausible poses and shapes such as unnatural bends and physically impossible interpenetrations. The first step of SMPLify involves an optimization over the camera translation and body orientation, while keeping the model pose and shape fixed.  After estimating the camera pose, SMPLify attempts to minimize Eq. \ref{equ:overall_loss} with a four-stage fitting procedure via a quasi-newton method. However, this whole process is very slow and remains prone to local minima. It is noteworthy that, the update step size $\lambda$ is usually subject to either a manual schedule or obtained through a deterministic algorithm such as line search. That implies that the step sizes are always chosen according to a predefined routine.\\

\myparagraph{Learned Gradient Descent for Human Shape Recovery.}\\
While finding good solutions with existing iterative optimization algorithms is challenging, they \emph{can} produce very good registrations to unseen data \emph{if} bad local minima are avoided. This is partially due to the iterative refinement and the absence of overfitting that can be problematic for regression methods.  
Motivated by recent ideas that connect the optimization and learning communities \cite{adler2017solving,adler2018learned,flynn2019deepview,andrychowicz2016learning}, we propose to replace the standard gradient descent rule with a learned per-parameter update:
{
\begin{align}
\Theta_{n+1} = \Theta_n + \mathcal{N}_w(\frac{\partial \mathcal{L}(\Theta_n) }{\partial \Theta_n}, \Theta_n,x)
\label{equ:updaterule_new}
\end{align}
}%
where $\mathcal{N}_w$ is a deep network parameterized by a set of weights $w$. The network processes the gradients and the current state of the model parameters to generate an update. Notice that $\lambda$ and $\Phi(\Theta)$ have been merged into $\mathcal{N}_w$ to jointly learn parameter prior, regularization terms and gradient update rule. This enables $\mathcal{N}_w$ to generate adaptive, parameter-specific updates that allow the fitting algorithm to stay on
the manifold of natural poses and shape and to recover from local minima. Furthermore, this process allows the optimization method to take large steps and hence to converge more quickly.

The inference process is illustrated in Fig. \ref{fig:inference}. Given a monocular image as input, an off-the-shelf body joint detector is run to attain the estimated target 2D keypoints $x$.
For the first iteration, we initialize $\Theta_0$ with zero values. After applying the re-projection process via Eq. \ref{equ:projection}, the error between predicted and observed measurements $\mathcal{L}(\Theta_0) = L_{reproj}(\hat{x}_0, x)$ is calculated. We then compute the partial derivative of the loss wrt the model parameters $\frac{\partial \mathcal{L}(\Theta_0) }{\partial \Theta_0}$.
Together with the current state of $\Theta_0$ and target $x$, the gradient $\frac{\partial \mathcal{L}(\Theta_0) }{\partial \Theta_0}$ is passed into the Gradient Updating Network to obtain the update term $\Delta \Theta_0$. Finally, $\Theta_0$ is updated to $\Theta_1$ by adding $\Delta \Theta_0$. The whole process will continue until the last iteration N to attain the estimated $\Theta_N$. N is usually less than four iterations.
\\

\myparagraph{Training}\\
For training, the \textit{only} source of data we use is a large dataset
of 3D meshes of human bodies of varying shape and in different poses \cite{AMASS:2019}. The SMPL parameters are obtained by running MOSH \cite{loper2014mosh} on different MoCap datasets. In order to ensure a fair comparison with other methods, we use the same subset as reported in \cite{pavlakos2019expressive,kanazawa2018end}. 
Please refer to supplementary materials for details.

\begin{algorithm}
\caption{- Training scheme}
\label{alg:training}
\begin{algorithmic}
    \STATE $\theta_{gt}, \beta_{gt} \leftarrow $ sample from database \\
    \STATE $ R_{gt},t_{gt},s_{gt} \leftarrow $ randomly sample within feasible range\\
    \STATE $\Theta_{gt} \leftarrow\left \{\theta_{gt},\beta_{gt},R_{gt},t_{gt},s_{gt}  \right \} $ \\
    \STATE $X_{gt} \leftarrow W M(\theta_{gt},\beta_{gt}) $ \\
    \STATE $x_{gt} \leftarrow s_{gt}\Pi(R_{gt}X_{gt}(\theta_{gt},\beta_{gt})) + t_{gt}$ \\
    \STATE $\Theta_0 \leftarrow \left \{\theta_0,\beta_0,R_0,t_0,s_0  \right \} \leftarrow 0$
      \FOR{$n = 0,...,N-1$ }
           \STATE $X_{n} \leftarrow W M(\theta_{n},\beta_{n}) $ \\
          \STATE $\hat{x}_n \leftarrow s_n\Pi(R_nX_n(\theta_n,\beta_n)) + t_n$ \\
          \STATE $\mathcal{L}(\Theta_n) \leftarrow L_{reproj}(\hat{x}_n, x_{gt})$ \\
          \STATE $\Delta \Theta_n \leftarrow \mathcal{N}_w(\frac{\partial \mathcal{L}(\Theta_n) }{\partial \Theta_n}, \Theta_n, x_{gt})$
          \\
          \STATE $\Theta_{n+1} \leftarrow \Theta_{n} +\Delta \Theta_n$
          \STATE $L_{\Theta_n} \leftarrow || \Theta_n - \Theta_{gt} ||_1$
      \ENDFOR
\end{algorithmic}
\end{algorithm}

The training algorithm for a single sample is given in pseudo-code in Alg. \ref{alg:training} and illustrated in Fig. \ref{fig:pipeline}.
During training, we randomly sample the pose $\theta_{gt}$ and shape $\beta_{gt}$ pair from the dataset, and re-project them with randomly generated camera extrinsics in order to obtain corresponding ground-truth 2D keypoints $x_{gt}$.  For illustration purposes we unroll the N-iteration process in Fig. \ref{fig:pipeline}, obtaining the full training architecture. More specifically, for each batch of training samples, we initialize $\Theta_0$ with zero values. After re-projection Eq. \ref{equ:projection}, the error between predicted and ground-truth 2D measurements $\mathcal{L}(\Theta_0) =  L_{reproj}(\hat{x}_0, x_{gt})$ is calculated.  The partial derivative $\frac{\partial \mathcal{L}(\Theta_0) }{\partial \Theta_0}$ is then computed and fed as input to the Gradient Updating Network. The other inputs are the current state of $\Theta_0$ and the target 2D measurements $x$. Once the gradient update term $\Delta \Theta_0$ is obtained, $\Theta$ is updated to $\Theta_1$ by adding it to $\Theta_0$. The whole process will continue until convergence to get $\Theta_N$. 

\figureNetwork
The training loss is only based on the error between true $\Theta_{gt}$ and estimated $\Theta_n$. During training, in order to bridge the gap between perfect 2D joints $x_{gt}$ obtained via re-projection to noisy 2D detections with missing joints from the CNN-based off-the-shelf detector, we randomly dropout some joints of $x_{gt}$.
Importantly, we do not leverage any image data for training and the only form of supervision stems from MoCap data processed via MOSH~\cite{loper2014mosh}.

\newcommand{\tablePW}{
\begin{table}[h]
\centering
\begin{tabular}{lcc}
\hline
Method                                                     &  Image + 3D annotation  & Rec. Error \\ \hline

HMR (with additional 3D data) \cite{kanazawa2018end}                             &Yes      & 81.3       \\
Kanazawa \emph{et al.} \cite{kanazawa2019learning}      &Yes     & 72.6       \\
Arnab \emph{et al.} \cite{arnab2019exploiting}            &Yes   & 72.2       \\
Kolotorous \emph{et al.} \cite{kolotouros2019convolutional} &Yes & 70.2       \\
SPIN (with additional 3D data) \cite{kolotouros2019learning}      &Yes                     & 59.2       \\
\hline
SMPLify \cite{bogo2016keep}                               &No  & 106.1      \\
Ours                                                      &No   & \textbf{55.9}       \\ \hline
\end{tabular}
\caption{\textbf{Evaluation on 3DPW.} Mean reconstruction errors in mm. Our method outperforms SMPLify by a significant margin. We achieve the state-of-art performance even compared with image based methods that use additional  datasets with expensive image-to-3D annotations and 3D pseudo-labels in the case of \cite{kolotouros2019learning}.}
\label{table:PW}
\end{table}
}

\newcommand{\tableHuman}{
\begin{table}[h]
\centering
\begin{tabular}{lc}
\hline
Method                                           & Rec. Error \\ \hline
Lasssner \emph{et al.} \cite{lassner2017unite}   & 93.9       \\
SMPLify \cite{bogo2016keep}                      & 82.3       \\
SMPLify (with GT 2D) \cite{bogo2016keep}         & 71.1       \\
SMPLify-X (with SMPL Body model) \cite{pavlakos2019expressive}          & 75.9       \\
Pavlakos \emph{et al.} \cite{pavlakos2018learning} & 75.9     \\
HMR (unpaired) \cite{kanazawa2018end}            & 66.5       \\
SPIN (unpaired) \cite{kolotouros2019learning}    & 62.0       \\
HMR \cite{kanazawa2018end}            & 56.8       \\
SPIN \cite{kolotouros2019learning}    & 41.1       \\
Ours                                             & \textbf{56.4}       \\ \hline
\end{tabular}
\caption{\textbf{Evaluation on the Human 3.6M set with Protocol 2.} The numbers are mean reconstruction errors in mm.  We compare with approaches that output a mesh of the human body. Our approach achieves the state-of-the-art performance.}
\label{table:Human}
\end{table}
}

\newcommand{\tableHumannew}{
\begin{table}[h]
\centering
\begin{tabular}{lcc}
\hline
Method                      &  Image + 3D annotation                     & Rec. Error \\ \hline
Lasssner \emph{et al.} \cite{lassner2017unite} &Yes  & 93.9       \\

Pavlakos \emph{et al.} \cite{pavlakos2018learning} &Yes & 75.9     \\

NBF  \cite{omran2018neural}  &Yes  & 59.9       \\
HMR (with additional 3D data)\cite{kanazawa2018end}       &Yes     & 56.8       \\
SPIN (with additional 3D data)\cite{kolotouros2019learning}  &Yes  & 41.1       \\
\hline
SMPLify \cite{bogo2016keep}           &No           & 82.3       \\
SMPLify-X (with SMPL Body model) \cite{pavlakos2019expressive}        &No  & 75.9       \\
SMPLify (with GT 2D) \cite{bogo2016keep}   &No      & 71.1       \\
HMR  \cite{kanazawa2018end}      &No      & 66.5       \\
SPIN  \cite{kolotouros2019learning}  &No  & 62.0       \\

Ours                               &No              & \textbf{56.4}       \\ \hline
\end{tabular}
\caption{\textbf{Evaluation on H3.6M.} Mean reconstruction errors in mm. We compare with approaches that output a mesh of the human body. The lower half of the table contains methods that do not require image-to-3D annotations (such as ours). We achieve state-of-the-art performance.}
\label{table:Human}
\end{table}
}

\newcommand{\tableEHF}{
\begin{table}[h]
\centering
\caption{\textbf{Evaluation on EHF dataset.} Vertex-to-vertex mean reconstruction errors in mm. Our method consistently outperforms different versions of SMPLify. No image based methods have conducted experiment on this dataset.}
\begin{tabular}{lc}
\hline
Method                                                      & v-v \\ \hline
SMPLify \cite{bogo2016keep}                                 & 73.8     \\
SMPLify-X (with SMPL Body model) \cite{pavlakos2019expressive}                     & 57.6      \\
Ours                                                        & \textbf{54.7}       \\ \hline
\end{tabular}
\label{table:EHF}
\end{table}
}

\newcommand{\tableComponents}{
\begin{table}[h]
\centering
\caption{\textbf{Ablation on gradient components}. Each entry represents an experiment including the gradient components labeled as target 2D pose ($x$), current estimated $\Theta$ ($\hat{\Theta}$), and true gradient ($\nabla\hat{\Theta}$). Runs are sorted in order of descending reconstruction error (mm) on Human 3.6M test set.}
\begin{tabular}{lc}
\hline
Input                                      & Rec. Error \\ \hline
$x$ \space \space    - \space \space \space      -                                        & 67.7       \\
$x$ \space  \space  - \space  $\nabla\hat{\Theta}$                      & 66.2      \\
$x$ \space   $\hat{\Theta}$ \space  \space   -                            & 62.9       \\
- \space \space    $\hat{\Theta}$ $\nabla\hat{\Theta}$           & 62.5       \\
$x$ \space $\hat{\Theta}$ $\nabla\hat{\Theta}$         & 59.2      \\ \hline
\end{tabular}
\label{table:components_old}
\end{table}
}

\newcommand{\tableSteps}{
\begin{table}[h]
\centering
\caption{\textbf{Ablation on number of updating iterations.} We also measure the effect of varying the number of updating iterations from 1 to 4 for training.}
\begin{tabular}{lc}
\hline
\#Iterations                                      & Rec. Error \\ \hline
1                                        & 66.3       \\
2                            & 62.1       \\
3                      & 59.2      \\

4       & 59.2      \\ \hline
\end{tabular}
\label{table:iteration_old}
\end{table}
}

\newcommand{\tableablation}{
\begin{minipage}{0.95\textwidth}
    \begin{minipage}[b]{0.45\textwidth}
        \captionof{table}{\textbf{Ablation on gradient components}.  Each entry represents an experiment including the gradient components labeled as target 2D pose ($x$), current estimated $\hat{\Theta}$, and unmapped gradient ($\nabla\hat{\Theta}$).}
        \centering
        \begin{tabular}{lc}
\hline
Input                                      & Rec. Error \\ \hline
$x$ \space \space    - \space \space \space      -                                        & 67.7       \\
$x$ \space  \space  - \space  $\nabla\hat{\Theta}$                      & 64.2      \\
$x$ \space   $\hat{\Theta}$ \space  \space   -                            & 61.9       \\
- \space \space    $\hat{\Theta}$ $\nabla\hat{\Theta}$           & 60.5       \\
$x$ \space $\hat{\Theta}$ $\nabla\hat{\Theta}$         & 56.4      \\ \hline
\end{tabular}

        \label{table:components}
    \end{minipage}
    \hfill
    \begin{minipage}[b]{0.45\textwidth}
        \centering
        \captionof{table}{\textbf{Ablation on number of iterations.} We also measure the effect of varying the number of iterations from 1 to 5 for training. The optimization converges around 4 iterations on average.}
        \begin{tabular}{lc}
\hline
\#Iterations                                      & Rec. Error \\ \hline
1                                        & 66.3       \\
2                            & 62.1       \\
3                      & 57.2      \\

4       & 56.6      \\ 
5      & 56.4      \\ \hline
\end{tabular}
        \label{table:iteration}
    \end{minipage}
\end{minipage}
}

\newcommand{\tableComponentstest}{
\begin{table}[h]
\centering
\begin{tabular}{lccc}
\hline
Method                                      & Rec. Error \\ \hline
$x,&\hat{\Theta},&$\nabla\hat{\Theta}$         & 59.2      \\ \hline
\end{tabular}
\caption{\textbf{Ablation on gradient components}. Each entry represents an experiment including the gradient components labeled as target 2D pose ($x$), current estimated $\Theta$ ($\hat{\Theta}$), and true gradient ($\nabla\hat{\Theta}$). Runs are sorted in order of descending reconstruction error (mm) on Human 3.6M test set.}
\label{table:components}
\end{table}
}

\newcommand{\tableData}{
\begin{table}[]
\centering
\footnotesize
\begin{threeparttable}[]
\begin{tabular}{c:ccc:cc}
         & \multicolumn{3}{c:}{Training data}                                                                                                  & \multicolumn{2}{c}{MPJPE} \\ 
         & \thead{ Same \\ MoCap\tnote{1}}                   & \thead{Images\\+2D pose}                                                                         &  \thead{Images\\+3D pose}                      & \textcolor{black}{3DPW}\tnote{2}        & H3.6M          \\ \Xhline{2\arrayrulewidth}
{~\textsuperscript{*}NBF}  & no                      & H3.6M                                                                               & H3.6M                           & -         & 59.9      \\ \hdashline
{~\textsuperscript{*}HMR paired}   & \multirow{2}{*}{yes} & \multirow{2}{*}{\begin{tabular}[c]{@{}c@{}}6 datasets\tnote{3}\end{tabular}} & \multirow{2}{*}{\begin{tabular}[c]{@{}c@{}}2 datasets\tnote{4}\end{tabular}} & 81.3         & 56.8       \\
{~\textsuperscript{*}SPIN paired}  &                      &                                                                                &                            & 59.2         & \textbf{41.1}      \\ \hline \hline

{HMR unpaired}  & \multirow{2}{*}{yes} & \multirow{2}{*}{\begin{tabular}[c]{@{}c@{}}6 datasets\tnote{3}\end{tabular}} & \multirow{2}{*}{-}         & -            & 66.5       \\
{SPIN unpaired} &                      &                                                                                &                            & -            & 62         \\  \hdashline
SMPLify & \multirow{2}{*}{yes} & \multirow{2}{*}{-\tnote{5}}                                                             & \multirow{2}{*}{-}         & 106.1         & 82.3        \\
Ours &                      &                                                                                &                            & \textbf{57.7}        & \textbf{59.2}      \\ \Xhline{2\arrayrulewidth}

\end{tabular}
\begin{tablenotes} []
     \item[*] \footnotesize\textcolor{black}{Methods that require expensive aligned 3D pose annotations.}
     \item[1] \footnotesize{\textcolor{black}{We use identical MoCap data to the one in HMR and SPIN.}}
     \item[2] \footnotesize{\textcolor{black}{Challenging pure hold-out test set for all methods.}}
     \item[3] \footnotesize{In total \textbf{5.1M} samples from MS COCO(\textbf{200K}), LSP(\textbf{2K}), LSP-extended(\textbf{10K}), MPII(\textbf{25K}), H3.6M(\textbf{3.6M}), MPI-3DHP(\textbf{1.3M})} 
     \item[4] \footnotesize{In total \textbf{4.9M} samples from H3.6M(\textbf{3.6M}), MPI-3DHP(\textbf{1.3M})}
     \item[5] \footnotesize{\textcolor{black}{Beyond those used for training the standalone 2D pose detector.}}
  \end{tablenotes}
\end{threeparttable}
\caption{\textbf{Data usage}. 
}
\label{table:data}
\end{table}
}

\newcommand{\tableDatanew}{
\begin{table}[]
\centering
\footnotesize
\begin{threeparttable}[]
\begin{tabular}{c:ccc}
      \diagbox{Methods}{Training data}                                                                                              
         & \thead{ Same \\ MoCap\tnote{1}}                   & \thead{Images+2D pose}                                                                         &  \thead{Images+3D pose}                          \\ \Xhline{2\arrayrulewidth}

{~\textsuperscript{*}HMR (with additional 3D data)\cite{kanazawa2018end}}   & \multirow{2}{*}{yes} & \multirow{2}{*}{\begin{tabular}[c]{@{}c@{}}6 datasets\tnote{2}\end{tabular}} & \multirow{2}{*}{\begin{tabular}[c]{@{}c@{}}2 datasets\tnote{3}\end{tabular}}       \\
{~\textsuperscript{*}SPIN (with additional 3D data)\cite{kolotouros2019learning}}  &                      &                                                                                &                            \\ \hline \hline
{HMR \cite{kanazawa2018end}}  & \multirow{2}{*}{yes} & \multirow{2}{*}{\begin{tabular}[c]{@{}c@{}}6 datasets\tnote{2}\end{tabular}} & \multirow{2}{*}{-}              \\
{SPIN \cite{kolotouros2019learning}} &                      &                                                                                &                              \\  \hdashline
SMPLify \cite{bogo2016keep}& \multirow{2}{*}{yes} & \multirow{2}{*}{-\tnote{4}}                                                             & \multirow{2}{*}{-}         \\
Ours &                      &                                                                                &                              \\ \Xhline{2\arrayrulewidth}
\end{tabular}

\begin{tablenotes} []
     \item[*] \footnotesize\textcolor{black}{Methods that require expensive aligned 3D pose annotations.}
     \item[1] \footnotesize{\textcolor{black}{We use identical MoCap data to the one in HMR and SPIN.}}
     \item[2] \footnotesize{In total \textbf{5.1M} samples from MS COCO(\textbf{200K}), LSP(\textbf{2K}), LSP-extended(\textbf{10K}), MPII(\textbf{25K}), H3.6M(\textbf{3.6M}), MPI-3DHP(\textbf{1.3M})} 
     \item[3] \footnotesize{In total \textbf{4.9M} samples from H3.6M(\textbf{3.6M}), MPI-3DHP(\textbf{1.3M})}
     \item[4] \footnotesize{\textcolor{black}{Beyond those used for training the standalone 2D pose detector.}}
  \end{tablenotes}
\end{threeparttable}
\caption{\textbf{Data usage}. HMR and SPIN require the same MoCap data as ours and large amounts of additional  annotated  data, especially 4.9 M datasets with 3D annotation. 
}
\label{table:datanew}
\end{table}
}

\newcommand{\tableDatanewtest}{
\begin{table}[]
\centering
\footnotesize
\begin{threeparttable}[]
\begin{tabular}{c:ccc}
         & \multicolumn{3}{c:}{Training data}       \\                                                                                            
         & \thead{ Same \\ MoCap\tnote{1}}                   & \thead{Images\\+2D pose}                                                                         &  \thead{Images\\+3D pose}                          \\ \Xhline{2\arrayrulewidth}

{~\textsuperscript{*}HMR (with additional 3D data)\cite{kanazawa2018end}}   & \multirow{2}{*}{yes} & \multirow{2}{*}{\begin{tabular}[c]{@{}c@{}}6 datasets\tnote{2}\end{tabular}} & \multirow{2}{*}{\begin{tabular}[c]{@{}c@{}}\footnotesize{H3.6M(\textbf{3.6M}), MPI-3DHP(\textbf{1.3M})}\end{tabular}}       \\
{~\textsuperscript{*}SPIN (with additional 3D data)\cite{kolotouros2019learning}}  &                      &                                                                                &                            \\ \hline \hline
{HMR \cite{kanazawa2018end}}  & \multirow{2}{*}{yes} & \multirow{2}{*}{\begin{tabular}[c]{@{}c@{}}6 datasets\tnote{2}\end{tabular}} & \multirow{2}{*}{-}              \\
{SPIN \cite{kolotouros2019learning}} &                      &                                                                                &                              \\  \hdashline
SMPLify \cite{bogo2016keep}& \multirow{2}{*}{yes} & \multirow{2}{*}{-\tnote{4}}                                                             & \multirow{2}{*}{-}         \\
Ours &                      &                                                                                &                              \\ \Xhline{2\arrayrulewidth}
\end{tabular}

\begin{tablenotes} []
     \item[*] \footnotesize\textcolor{black}{Methods that require expensive aligned 3D pose annotations.}
     \item[1] \footnotesize{\textcolor{black}{We use identical MoCap data to the one in HMR and SPIN.}}
     \item[2] \footnotesize{In total \textbf{5.1M} samples from MS COCO(\textbf{200K}), LSP(\textbf{2K}), LSP-extended(\textbf{10K}), MPII(\textbf{25K}), H3.6M(\textbf{3.6M}), MPI-3DHP(\textbf{1.3M})} 
     \item[3] \footnotesize{In total \textbf{4.9M} samples from H3.6M(\textbf{3.6M}), MPI-3DHP(\textbf{1.3M})}
     \item[4] \footnotesize{\textcolor{black}{Beyond those used for training the standalone 2D pose detector.}}
  \end{tablenotes}
\end{threeparttable}
\caption{\textbf{Data usage}. Direct regression methods (HMR, SPIN) require the same MoCap data as ours and large amounts of additional  annotated  data  (6  datasets  with  2D  annotation  and 2 datasets with 3D annotation). 
}
\label{table:datanew}
\end{table}
}

\section{Experiments}
\subsection{Training Data Comparison} For training, we \textit{only} use a dataset of 3D meshes of human bodies of varying shape and in different poses \cite{AMASS:2019}. For clarity, we emphasize that our primary goal is to boost the robustness and effectiveness of iterative optimization due to their importance to register SMPL meshes to new, entirely unseen data (e.g., to produce new datasets). Therefore, the most related method is SMPLify \cite{bogo2016keep}. SMPLify leverages a learned prior and uses exactly the same MoCap data to train it. Thus, a direct comparison is fair. Note that our method can additionally be used as a standalone pose estimator and hence we compare to learning-based regression methods. All learning methods require stronger 3D supervision and much more data. For example, the current state-of-art approaches \cite{kanazawa2018end,kolotouros2019learning} utilize 6 additional datasets with 2D pose annotation (5.1M samples in total) and 2 additional datasets (4.9M samples in total) with 3D annotation. Since neither SMPLify nor ours leverages any image data at training or inference time, these can only be compared directly to methods that do not require the strongest form of 3D supervision: paired image-to-3D annotations. We list the performance of such methods only for completeness. 

\subsection{Test Datasets}
\label{sec:datasets}
\myparagraph{Human 3.6M \cite{ionescu2013human3}}
 is an indoor dataset. The 3D poses are obtained from a MoCap system. The dataset covers 7 subjects, each engaging in various activities such as walking, photo taking and dog walking. Following the evaluation protocol, we evaluate 1 of every 5 frames and only on the frontal camera (camera 3). We report the reconstruction error (MPJPE after procrustes alignment). Similar to other optimization based methods, we obtain 2D pose estimates from a CNN based pose detector (a stacked hourglass network \cite{newell2016stacked} trained on MPI \cite{andriluka20142d} and fine-tuned on the Human 3.6M training set).

\myparagraph{3DPW \cite{von2018recovering}}
is a challenging outdoor dataset which provides 3D pose and shape ground-truth obtained by fusing information of IMU and 2D keypoint detections. It covers complex natural poses in-the-wild and is currently the best benchmark for real-world performance.  Note that this dataset is \emph{not} used for training and a pure hold-out test set for all methods. We report the 3D pose reconstruction error on the test set. Frames in which less than 6 joints are detected are discarded as in \cite{kanazawa2019learning}. We use the 2D keypoints included in the dataset (from openpose~\cite{cao2018openpose}).

\myparagraph{EHF \cite{pavlakos2019expressive}}
The expressive hands and faces dataset (EHF) has 100 images with vertex level annotation of the whole human body. They are obtained first by fitting the SMPL-X model and then manually curated by an expert annotator. The dataset can be considered as dense pseudo ground-truth, according to its alignment quality. The pseudo ground-truth meshes allow to use a stricter vertex to-vertex (v2v) error metric, in contrast to the common paradigm of reporting 3D joint error, which does neither capture misalignment of the surface itself nor rotations along the bones. To the best of our knowledge, EHF is the only available real-world dataset with direct shape measurements (from 3D scans).\\

\tableablation
\subsection{Ablation and Iteration Studies}
\myparagraph{Inputs to the Gradient Update Network}
In this experiment, we evaluate the importance of each of the input components for the Gradient Update Network. To this end we discard one or more of the components and measure the influence in terms of final accuracy. The experiment is conducted on the Human 3.6M test set.
Results are shown in Tab. \ref{table:components}. When using only the target 2D pose $x$ as input, the network is equivalent to a residual network that operates by directly lifting $\Theta$ from 2D keypoints. As expected, in this configuration the model performs poorly since no prior knowledge on the human model is incorporated. Adding additional inputs gradually increases performance with the best performance is achieved when all gradient update components are used. In the following experiments, we fix our setting to use all three input components. \\

\myparagraph{Number of iterations during training}
In Tab. \ref{table:iteration} we measure the effect of varying the number of update iterations from 1 to 5 for each training sample. The experiment is also conducted on the Human 3.6M test set. Not surprisingly, the results improve as the number of iterations increases. We note that after four iterations there is no further improvement (we use four iterations in our experiments). Note that extrapolation beyond this training window is possible within reason (cf. Figure \ref{fig:comparison_smpl}).

\myparagraph{Comparison with direct lifting from 2D pose}
We also compare to the baseline method proposed in \cite{martinez2017simple}, i.e., directly lifting 2D pose to 3D pose. The main difference is that direct lifting is non-iterative and does not take gradients as input. Its reconstruction error is 67.8, which is $15\%$ worse than ours. This suggests that gradient-based refinement indeed leads to better and more detailed registration compared to simple lifting.

\subsection{Comparison with Other Methods}
We compare results with other state-of-art methods on the three datasets as introduced in Sec. \ref{sec:datasets}.
\tableHumannew

\myparagraph{Human 3.6M \cite{ionescu2013human3}}
We report the reconstruction error on Protocol 2 following \cite{bogo2016keep}. Since our method does not use images for 3D fitting, it is only possible to directly compare to  methods which output SMPL parameters from a single image and \emph{do not} require image-to-3D paired data for training.
As shown in Tab. \ref{table:Human}, we outperform SMPLify \cite{bogo2016keep} by a large margin, even when they fit to ground truth 2D keypoints. Fig. \ref{fig:comparison_smpl} shows two representative examples of the progression of the reconstruction error as function of iterations. Compared to SMPLify, \emph{ours} converges much quicker and more stable. The images in the inset provide insights into why this is the case. Our approach jointly manipulates global and body pose and arrives at a good solution in few steps. Note that the steps in light blue indicate steps beyond the training window (N = 4). While the SMPLify (red) shows clear signs of the adaptive weighting of different loss terms, which can also be seen in the intermediate states, where first the global pose is adjusted and body pose is only optimized later (cf. Fig. \ref{fig:comparison_smpl}, insets). Fig. \ref{fig:comparison_smpl}, right illustrates a case where SMPLify gets stuck in a bad local minima and fails to recover from it.

Ours also achieves a lower reconstruction error than state-of-the-art regression methods \cite{kanazawa2018end,kolotouros2019learning} if compared fairly to ours. That is we directly compare the setting in which these methods do not use image-to-3D paired information since our method does not have access to this additional data. We use the same 3D MoCap data that \cite{kanazawa2018end,kolotouros2019learning} use to train their pose priors. While we outperform HMR even in the paired setting, SPIN performs better when allowed to use additional data. However, we note that H3.6M is a controlled and relatively small dataset and is known to be prone to overfitting. Qualitative results shown in Fig. \ref{fig:compare_new} show that regression methods are sensitive to scene clutter close to the person. Our results align better after iterative refinement. 

\figureComparisonSMPL
\tablePW
\myparagraph{3DPW \cite{von2018recovering}}
This in-the-wild dataset is used solely for testing, hence a better indicator of real-world performance. While direct regression methods (HMR, SPIN) use the same MoCap data as ours \emph{and} large amounts of additional  annotated  data, these methods exhibit a clear performance decrease on this dataset compared to H3.6M (see Tab. \ref{table:PW}). In this challenging setting, we significantly outperform optimization methods that use the same data as ours (45\% over SMPLify \cite{bogo2016keep}). Furthermore, our method outperforms regression  based  methods  (e.g., by  29\%  over HMR \cite{kanazawa2018end}), even in the setting where they are allowed to use additional 3D supervision. We also outperform the current state-of-art \cite{kolotouros2019learning}, which not only employs several datasets with image-to-3D labels but also leverages SMPLify to obtain pseudo 3D labels on a large scale dataset with 2D annotations.

\setlength{\intextsep}{0pt}
\begin{wraptable}{r}{4.7cm}
\vskip -0mm
\caption{\textbf{EHF dataset.} Vertex-to-vertex mean reconstruction errors in mm. Our method consistently outperforms different versions of SMPLify.
}
\label{tab:eval_gazemap}
\centering
\renewcommand*{\arraystretch}{1.1}
\begin{tabular}{lc}
\hline
Method                                                      & v-v \\ \hline
SMPLify \cite{bogo2016keep}                                 & 73.8     \\
SMPLify-X \cite{pavlakos2019expressive}                     & 57.6      \\
Ours                                                        & \textbf{54.7}       \\ \hline
\end{tabular}
\end{wraptable}
\myparagraph{EHF \cite{pavlakos2019expressive}}
Finally, we also report results on the EHF dataset for detailed vertex level evaluation. To the best of our knowledge, EHF is the only available real-world dataset with direct shape measurement (from 3D scans).
We achieve 54.7mm error in terms of vertex-to-vertex comparison with ground truth, which consistently outperforms different versions of SMPLify \cite{bogo2016keep,pavlakos2019expressive}. To be noticed, no regression based methods have reported performance on detailed vertex level evaluation.

\subsection{Qualitative Results}
Fig. \ref{fig:compare_new} shows comparisons to regression based methods. The highlight depicts instances where iterative refinement (ours) better aligns details such as the lower limbs. 
Fig. \ref{fig:collage} shows more qualitative results of our approach from different datasets, demonstrating consistent behavior irrespective of the dataset. 
 
 \subsection{Speed and Model Size.}
 On the same hardware, our method converges on average in 120 \textbf{ms}, whereas SMPLify generally takes 1 to 2 minutes. This 500x speed-up is due to: \textbf{i)} fewer iterations (5 vs. 100) and \textbf{ii)} requiring only first order derivatives while SMPLify relies on second order methods.  Our gradient updating network is lightweight, with 4.3M parameters and 8M FLOPs (0.008G).

 \figureCompare
 \figureCollage
 
\section{Conclusion}
We propose a novel optimization algorithm for  3D  human  body model fitting.   We  replace  the  normal  gradient descent update rule, which depends on a hand-tuned step-size, with a deep network to predict per-parameter, state dependent updates. Our method significantly improves over the closest state-of-art, formed by non-convex optimization methods such as SMPLify, in terms of convergence, speed,  and accuracy when using the same MoCap data. Accurately annotating 3D human shape in unconstrained natural environments is extremely challenging and will remain so for a long time. Hence, improved optimization based methods have the potential to provide more training data by fitting 2D annotated images. 

\noindent\textbf{Acknowledgement.} This research was partially supported by the Max Planck ETH Center for Learning Systems and a research gift from NVIDIA.

\bibliographystyle{splncs04}
\bibliography{egbib}

\begin{thebibliography}{10}
\providecommand{\url}[1]{\texttt{#1}}
\providecommand{\urlprefix}{URL }
\providecommand{\doi}[1]{https://doi.org/#1}

\bibitem{adler2017solving}
Adler, J., {\"O}ktem, O.: Solving ill-posed inverse problems using iterative
  deep neural networks. Inverse Problems  \textbf{33}(12),  124007 (2017)

\bibitem{adler2018learned}
Adler, J., {\"O}ktem, O.: Learned primal-dual reconstruction. IEEE transactions
  on medical imaging  \textbf{37}(6),  1322--1332 (2018)

\bibitem{andriluka20142d}
Andriluka, M., Pishchulin, L., Gehler, P., Schiele, B.: 2d human pose
  estimation: New benchmark and state of the art analysis. In: Proceedings of
  the IEEE Conference on computer Vision and Pattern Recognition. pp.
  3686--3693 (2014)

\bibitem{andrychowicz2016learning}
Andrychowicz, M., Denil, M., Gomez, S., Hoffman, M.W., Pfau, D., Schaul, T.,
  Shillingford, B., De~Freitas, N.: Learning to learn by gradient descent by
  gradient descent. In: Advances in neural information processing systems. pp.
  3981--3989 (2016)

\bibitem{anguelov2005scape}
Anguelov, D., Srinivasan, P., Koller, D., Thrun, S., Rodgers, J., Davis, J.:
  Scape: shape completion and animation of people. In: ACM transactions on
  graphics (TOG). vol.~24, pp. 408--416. ACM (2005)

\bibitem{arnab2019exploiting}
Arnab, A., Doersch, C., Zisserman, A.: Exploiting temporal context for 3d human
  pose estimation in the wild. In: Proceedings of the IEEE Conference on
  Computer Vision and Pattern Recognition. pp. 3395--3404 (2019)

\bibitem{bogo2016keep}
Bogo, F., Kanazawa, A., Lassner, C., Gehler, P., Romero, J., Black, M.J.: Keep
  it smpl: Automatic estimation of 3d human pose and shape from a single image.
  In: European Conference on Computer Vision. pp. 561--578. Springer (2016)

\bibitem{cao2018openpose}
Cao, Z., Hidalgo, G., Simon, T., Wei, S.E., Sheikh, Y.: Openpose: realtime
  multi-person 2d pose estimation using part affinity fields. arXiv preprint
  arXiv:1812.08008  (2018)

\bibitem{flynn2019deepview}
Flynn, J., Broxton, M., Debevec, P., DuVall, M., Fyffe, G., Overbeck, R.,
  Snavely, N., Tucker, R.: Deepview: View synthesis with learned gradient
  descent. arXiv preprint arXiv:1906.07316  (2019)

\bibitem{guan2009estimating}
Guan, P., Weiss, A., Balan, A.O., Black, M.J.: Estimating human shape and pose
  from a single image. In: 2009 IEEE 12th International Conference on Computer
  Vision. pp. 1381--1388. IEEE (2009)

\bibitem{guler2019holopose}
Guler, R.A., Kokkinos, I.: Holopose: Holistic 3d human reconstruction
  in-the-wild. In: Proceedings of the IEEE Conference on Computer Vision and
  Pattern Recognition. pp. 10884--10894 (2019)

\bibitem{hadamard1902problemes}
Hadamard, J.: Sur les probl{\`e}mes aux d{\'e}riv{\'e}es partielles et leur
  signification physique. Princeton university bulletin pp. 49--52 (1902)

\bibitem{hasler2010multilinear}
Hasler, N., Ackermann, H., Rosenhahn, B., Thorm{\"a}hlen, T., Seidel, H.P.:
  Multilinear pose and body shape estimation of dressed subjects from image
  sets. In: 2010 IEEE Computer Society Conference on Computer Vision and
  Pattern Recognition. pp. 1823--1830. IEEE (2010)

\bibitem{ionescu2013human3}
Ionescu, C., Papava, D., Olaru, V., Sminchisescu, C.: Human3. 6m: Large scale
  datasets and predictive methods for 3d human sensing in natural environments.
  IEEE transactions on pattern analysis and machine intelligence
  \textbf{36}(7),  1325--1339 (2013)

\bibitem{joo2018total}
Joo, H., Simon, T., Sheikh, Y.: Total capture: A 3d deformation model for
  tracking faces, hands, and bodies. In: Proceedings of the IEEE Conference on
  Computer Vision and Pattern Recognition. pp. 8320--8329 (2018)

\bibitem{kanazawa2018end}
Kanazawa, A., Black, M.J., Jacobs, D.W., Malik, J.: End-to-end recovery of
  human shape and pose. In: Proceedings of the IEEE Conference on Computer
  Vision and Pattern Recognition. pp. 7122--7131 (2018)

\bibitem{kanazawa2019learning}
Kanazawa, A., Zhang, J.Y., Felsen, P., Malik, J.: Learning 3d human dynamics
  from video. In: Proceedings of the IEEE Conference on Computer Vision and
  Pattern Recognition. pp. 5614--5623 (2019)

\bibitem{kolotouros2019learning}
Kolotouros, N., Pavlakos, G., Black, M.J., Daniilidis, K.: Learning to
  reconstruct 3d human pose and shape via model-fitting in the loop. In:
  Proceedings of the IEEE International Conference on Computer Vision. pp.
  2252--2261 (2019)

\bibitem{kolotouros2019convolutional}
Kolotouros, N., Pavlakos, G., Daniilidis, K.: Convolutional mesh regression for
  single-image human shape reconstruction. In: Proceedings of the IEEE
  Conference on Computer Vision and Pattern Recognition. pp. 4501--4510 (2019)

\bibitem{lassner2017unite}
Lassner, C., Romero, J., Kiefel, M., Bogo, F., Black, M.J., Gehler, P.V.: Unite
  the people: Closing the loop between 3d and 2d human representations. In:
  Proceedings of the IEEE Conference on Computer Vision and Pattern
  Recognition. pp. 6050--6059 (2017)

\bibitem{loper2014mosh}
Loper, M., Mahmood, N., Black, M.J.: Mosh: Motion and shape capture from sparse
  markers. ACM Transactions on Graphics (TOG)  \textbf{33}(6), ~220 (2014)

\bibitem{loper2015smpl}
Loper, M., Mahmood, N., Romero, J., Pons-Moll, G., Black, M.J.: Smpl: A skinned
  multi-person linear model. ACM transactions on graphics (TOG)
  \textbf{34}(6), ~248 (2015)

\bibitem{AMASS:2019}
Mahmood, N., Ghorbani, N., F.~Troje, N., Pons-Moll, G., Black, M.J.: Amass:
  Archive of motion capture as surface shapes. In: The IEEE International
  Conference on Computer Vision (ICCV) (Oct 2019),
  \url{https://amass.is.tue.mpg.de}

\bibitem{von2018recovering}
von Marcard, T., Henschel, R., Black, M.J., Rosenhahn, B., Pons-Moll, G.:
  Recovering accurate 3d human pose in the wild using imus and a moving camera.
  In: Proceedings of the European Conference on Computer Vision (ECCV). pp.
  601--617 (2018)

\bibitem{martinez2017simple}
Martinez, J., Hossain, R., Romero, J., Little, J.J.: A simple yet effective
  baseline for 3d human pose estimation. In: Proceedings of the IEEE
  International Conference on Computer Vision. pp. 2640--2649 (2017)

\bibitem{mehta2017vnect}
Mehta, D., Sridhar, S., Sotnychenko, O., Rhodin, H., Shafiei, M., Seidel, H.P.,
  Xu, W., Casas, D., Theobalt, C.: Vnect: Real-time 3d human pose estimation
  with a single rgb camera. ACM Transactions on Graphics (TOG)  \textbf{36}(4),
  ~44 (2017)

\bibitem{newell2016stacked}
Newell, A., Yang, K., Deng, J.: Stacked hourglass networks for human pose
  estimation. In: European conference on computer vision. pp. 483--499.
  Springer (2016)

\bibitem{omran2018neural}
Omran, M., Lassner, C., Pons-Moll, G., Gehler, P., Schiele, B.: Neural body
  fitting: Unifying deep learning and model based human pose and shape
  estimation. In: 2018 International Conference on 3D Vision (3DV). pp.
  484--494. IEEE (2018)

\bibitem{pavlakos2019expressive}
Pavlakos, G., Choutas, V., Ghorbani, N., Bolkart, T., Osman, A.A., Tzionas, D.,
  Black, M.J.: Expressive body capture: 3d hands, face, and body from a single
  image. In: Proceedings of the IEEE Conference on Computer Vision and Pattern
  Recognition. pp. 10975--10985 (2019)

\bibitem{pavlakos2018learning}
Pavlakos, G., Zhu, L., Zhou, X., Daniilidis, K.: Learning to estimate 3d human
  pose and shape from a single color image. In: Proceedings of the IEEE
  Conference on Computer Vision and Pattern Recognition. pp. 459--468 (2018)

\bibitem{pishchulin2016deepcut}
Pishchulin, L., Insafutdinov, E., Tang, S., Andres, B., Andriluka, M., Gehler,
  P.V., Schiele, B.: Deepcut: Joint subset partition and labeling for multi
  person pose estimation. In: Proceedings of the IEEE Conference on Computer
  Vision and Pattern Recognition. pp. 4929--4937 (2016)

\bibitem{sigal2008combined}
Sigal, L., Balan, A., Black, M.J.: Combined discriminative and generative
  articulated pose and non-rigid shape estimation. In: Advances in neural
  information processing systems. pp. 1337--1344 (2008)

\bibitem{sun2018integral}
Sun, X., Xiao, B., Wei, F., Liang, S., Wei, Y.: Integral human pose regression.
  In: Proceedings of the European Conference on Computer Vision (ECCV). pp.
  529--545 (2018)

\bibitem{tan2018indirect}
Tan, V., Budvytis, I., Cipolla, R.: Indirect deep structured learning for 3d
  human body shape and pose prediction  (2018)

\bibitem{tung2017self}
Tung, H.Y., Tung, H.W., Yumer, E., Fragkiadaki, K.: Self-supervised learning of
  motion capture. In: Advances in Neural Information Processing Systems. pp.
  5236--5246 (2017)

\bibitem{varol2018bodynet}
Varol, G., Ceylan, D., Russell, B., Yang, J., Yumer, E., Laptev, I., Schmid,
  C.: Bodynet: Volumetric inference of 3d human body shapes. In: Proceedings of
  the European Conference on Computer Vision (ECCV). pp. 20--36 (2018)

\bibitem{varol2017learning}
Varol, G., Romero, J., Martin, X., Mahmood, N., Black, M.J., Laptev, I.,
  Schmid, C.: Learning from synthetic humans. In: Proceedings of the IEEE
  Conference on Computer Vision and Pattern Recognition. pp. 109--117 (2017)

\bibitem{xiang2019monocular}
Xiang, D., Joo, H., Sheikh, Y.: Monocular total capture: Posing face, body, and
  hands in the wild. In: Proceedings of the IEEE Conference on Computer Vision
  and Pattern Recognition. pp. 10965--10974 (2019)

\bibitem{xu2019denserac}
Xu, Y., Zhu, S.C., Tung, T.: Denserac: Joint 3d pose and shape estimation by
  dense render-and-compare. In: Proceedings of the IEEE International
  Conference on Computer Vision. pp. 7760--7770 (2019)

\bibitem{zheng2019deephuman}
Zheng, Z., Yu, T., Wei, Y., Dai, Q., Liu, Y.: Deephuman: 3d human
  reconstruction from a single image. In: Proceedings of the IEEE International
  Conference on Computer Vision. pp. 7739--7749 (2019)

\bibitem{zhou2016sparseness}
Zhou, X., Zhu, M., Leonardos, S., Derpanis, K.G., Daniilidis, K.: Sparseness
  meets deepness: 3d human pose estimation from monocular video. In:
  Proceedings of the IEEE conference on computer vision and pattern
  recognition. pp. 4966--4975 (2016)

\end{thebibliography}
\end{document}